\documentclass{article}

\usepackage{microtype}
\usepackage{graphicx}
\usepackage{subfigure}
\usepackage{booktabs} 
\usepackage{url}
\usepackage{bbm}
\usepackage{amsmath}
\usepackage{times}
\usepackage{epsfig}
\usepackage{amsmath}
\usepackage{amssymb}
\usepackage{booktabs}
\usepackage{paralist}
\usepackage{multirow}
\usepackage{wrapfig}
\usepackage{hyperref}

\usepackage{color-edits}
\addauthor{pg}{red}
\addauthor{sg}{green}
\addauthor{ss}{orange}

\usepackage[accepted]{icml2020}

\icmltitlerunning{Exploiting Temporal Coherence for Multi-modal Video Categorization}

\begin{document}

\twocolumn[
\icmltitle{Exploiting Temporal Coherence for Multi-modal Video Categorization}



\icmlsetsymbol{equal}{*}

\begin{icmlauthorlist}
\icmlauthor{Palash Goyal}{equal,to}
\icmlauthor{Saurabh Sahu}{equal,to}
\icmlauthor{Shalini Ghosh}{equal,to}
\icmlauthor{Chul Lee}{to}
\end{icmlauthorlist}

\icmlaffiliation{to}{Visual Display Intelligence Lab, Samsung Research America}
\icmlcorrespondingauthor{Shalini Ghosh}{shalini.ghosh@samsung.com}

\icmlkeywords{Multi-modal modeling, temporal coherence, video categorization, transformer}

\vskip 0.3in
]



\printAffiliationsAndNotice{\icmlEqualContribution} 

\begin{abstract}
Multimodal ML models can process data in multiple modalities (e.g., video, images, audio, text) and are useful for video content analysis in a variety of problems (e.g., object detection, scene understanding). In this paper, we focus on the problem of video categorization by using a multimodal approach. We have developed a novel  temporal coherence-based regularization approach, which applies to different types of models (e.g., RNN, NetVLAD, Transformer). We demonstrate through experiments how our proposed multimodal video categorization models with temporal coherence out-perform strong state-of-the-art baseline models.
\end{abstract}

\section{Introduction and Motivation}
\label{sec:intro}

Multimodal machine learning (ML) models typically processes data from different modalities (e.g., video, image, audio, language) in order to do a task. For example, consider the task of identifying events in a video -- the ML model may be able to detect ``guitar" and "crowd" in the video frames, which along with loud music in the audio channel could enable the ML model to infer the event ``music concert". Multimodal ML models have some advantages: (a) can have more accuracy, since data from different modalities can reinforce each other or help in disambiguation; (b) can be more resilient to errors in the data, e.g., if the video channel goes dark due to a data glitch, close-caption text of relevant words combined with audio of the sound of gunfire, aeroplanes and explosions could help detect that the video is an action sequence from a war movie. 

These advantages can make multimodal ML models very suitable for different types of video content analysis  problems. In visual dialog, a ML model tries to understand a scene and have a conversation with a user about it --- using a multimodal ML model will help the system interact with the user along various modalities (e.g., speech, text), as well as process different aspects of a video (e.g., image, audio) while answering questions. In object detection, identifying an object from a video frame can be aided by supporting evidence from CC-text or audio. Finally, in content recommendation, a ML model can use multimodal AI to process the videos in a user's watch history and recommend new video content accordingly.
In this paper, we consider the content recommendation problem where the ML model suggests new video content to a user based on whether particular characteristics in that video matches the corresponding characteristics inferred from the sequence of watched videos in the user's history. In our discussion, the particular characteristics of the video that we will consider to do the match for content recommendation are fine-grained categories (e.g., Entertainment/Concert, Action/War). 

A key innovation we propose in this paper is efficient application of the principle of temporal coherence in different video analysis models.
Temporal coherence is a regularization technique that (a) helps ensure that the ML model processing the video has coherent behavior when processing frames that are  adjacent or within a small neighborhood~\cite{MobahiCW09}, (b) helps coming up with unsupervised representations of video frames where nearby frames have similar embeddings~\cite{Wang2015}. Our key contribution is showing how this temporal coherence principle can be efficiently incorporated into different types of models, specifically attention RNN~\cite{Bahdanau14}, Transformer~\cite{Vaswani17} and NetVLAD~\cite{Arandjelovic16}, albeit in different ways. In attention RNN or Transformer models the temporal coherence is applied to the attention variables, while in NetVLAD models temporal coherence is applied to the prediction vectors. We prove that temporal coherence can be efficiently enforced in different models using a convolutional layer used appropriately in the model. We also show empirically how incorporating temporal coherence helps NetVLAD, attention RNN and Transformer give better performance --- an ensemble of these 3 models with temporal coherence regularization out-performs strong baselines in the video content categorization problem.

Section~\ref{sec:formulation} discusses the main formulation of temporal coherence and how it can be applied to different models --- we outline TC-RNN, TC-NetV and TC-TM, which are temporally coherent variants of attention RNN, NetVLAD and Transformer models respectively. Section~\ref{sec:convolution} outlines how temporal coherence can be efficiently implemented in these three models using a convolutional and pooling layer. Section~\ref{sec:experiments} discusses the experimental setup and analyzes the results, especially showing how an ensemble of the 3 TC models gives state-of-the-art performance. Section~\ref{sec:related}  outlines related research, while Section~\ref{sec:conclusions}.
concludes the paper with a summary and discussion of possible future work.
\section{Temporal Coherence}
\label{sec:formulation}

In this section, we outline in detail how the temporal coherence principle is incorporated into different ML models, specifically attention RNN, NetVLAD and Transformer.

\subsection{Model output with Temporal Coherence}

One direct way to enforce temporal coherence in a model is to incorporate it into the model output as a regularizer. Let us consider a homogeneous segment of a video, which we define as a ``video segment" --- it corresponds to a video segment that is bounded on two sides (beginning and end) by scene transitions. Within that segment of video, the ML model will have a strong prior of having similar prediction labels in its output (e.g., similar objects, similar events, similar activities, depending on the prediction task). 

We can encode this domain knowledge by considering that frames in a segment have a strong prior of similar class predictions for the task at hand --- this can be encoded as a regularizer term in the loss function for the task. The overall loss function with the extra regularization term has the form: 
\begin{equation}
\label{eqn:overall_loss}
L_{overall} = L_{original} + \lambda L_{labelprior}
\end{equation}
where $L_{overall}$ is the final overall loss, $L_{original}$ is the original loss function, $L_{labelprior}$ is the label similarity loss corresponding to the label prior, and $\lambda$ is a hyper-parameter to help the model trade-off between these 2 loss terms. 


In our initial studies with different types of output regularizers, we realized that the label prior is often a strong condition to enforce, since it requires external knowledge of scene boundaries, and the results may be sensitive to choice of the hyper-parameter $\lambda$. So, we considered other methods of enforcing temporal coherence in video analysis models.

\subsection{NetVLAD with Temporal Coherence}

\begin{figure}[hbtp]
\centering
    \includegraphics[width=\columnwidth]{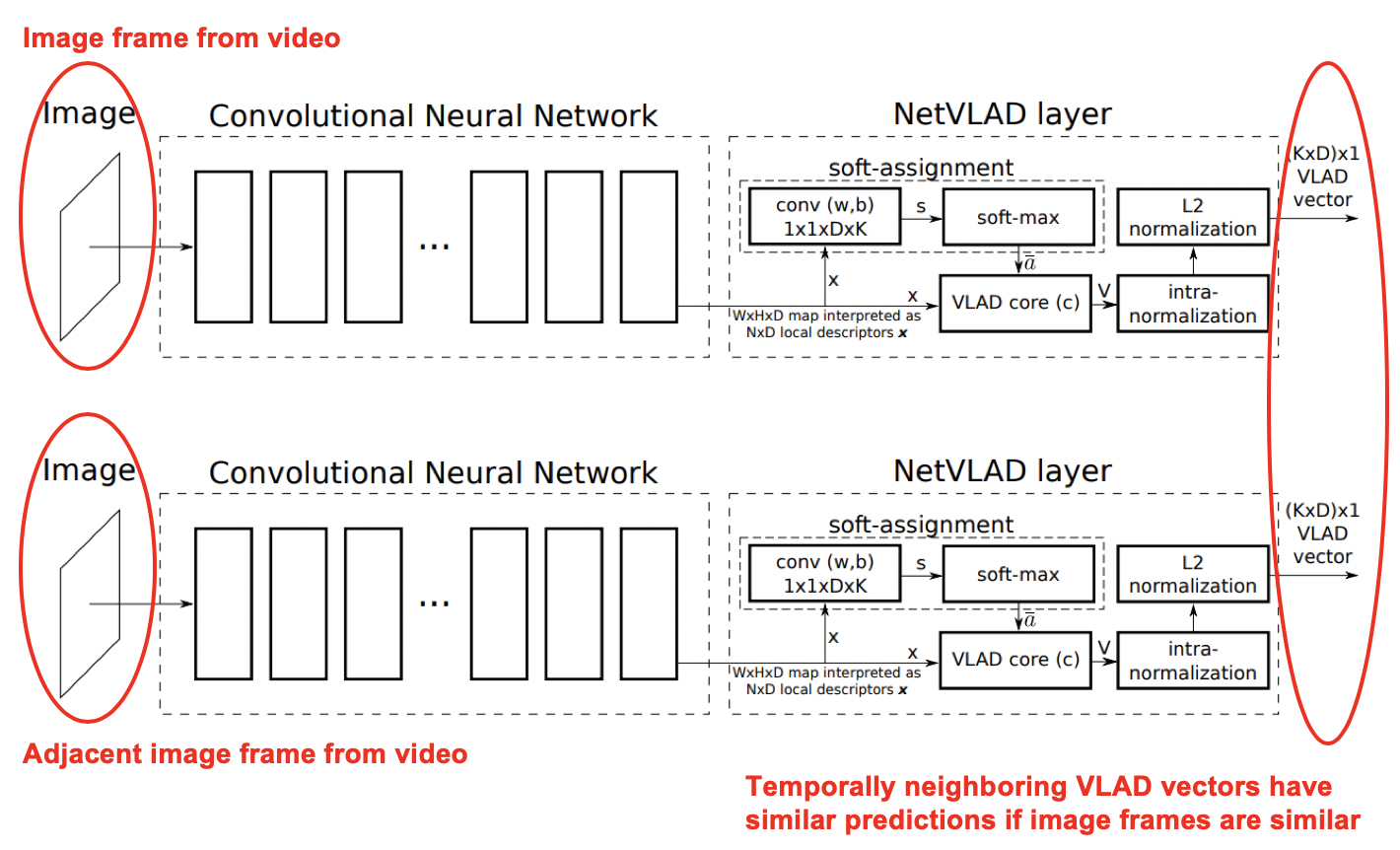}
    \caption{Temporal coherence principle applied to NetVLAD model~\cite{Arandjelovic16}.}
    \label{fig:netvlad-attention}
\end{figure}

Figure~\ref{fig:netvlad-attention} shows how the temporal coherence model would apply to the NetVLAD model~\cite{Arandjelovic16}.
In this case, the intuition is that if a frame has a particular set of soft assignment variables $\alpha_k(x_i)$ to the $K$ clusters (see Equation 3 in \cite{Arandjelovic16}), then the other frames in the same neighborhood have a high likelihood of having similar cluster assignment variable values since the salient descriptors in the frame have a high likelihood of being similar. This is especially true if the frames in the neighborhood are in the same scene in the video, i.e., the neighborhood does not cross a scene boundary --- if it does cross a scene boundary, the cluster assignment variables could change since the salient descriptors in the frame could change as the scene changes.

Let us consider the formulation in the original NetVLAD paper~\cite{Arandjelovic16}. We consider the following form of the assignment variable $\alpha_k$:
\[
\alpha_k(x_i) = \frac{\exp{(q_{ki})}}{\sum_{k'} \exp{(q_{k'i})}}, \mbox{where } q_{ki} = w_k^T x_i + b_k.
\]


In order to model the temporal coherence, we modify the soft assignment variable as follows:
\begin{equation}
\small
\alpha_k = \frac{\exp{(q_{ki})}.\exp\bigg(\sum_{j \in N_i} q_{kj}. \exp{-\big(||x_i-x_j||_2}\big)\bigg)}
{\sum_{k'} \exp{(q_{k'i})}.\exp\bigg(\sum_{j \in N_i} q_{k'j}. \exp{-\big(||x_i-x_j||_2}\big)\bigg)} 
\label{eqn:assign_coh}
\end{equation}
where $N_i$ is the neighborhood of node $i$ (say the nodes in the neighborhood $[i-L, i+L]$). 

Another variation of this formulation is to model the scene boundary, to capture the fact whether the frames in the same neighborhood will be temporally coherent or not. We consider $z$ to be the random variable indicating the {\em \it absence} of a scene boundary within a neighborhood --- it takes the value 1 when the scene boundary is absent and value 0 when the scene boundary is present. With this additional variable $z$, the updated equation is now:
\begin{equation}
\small
\alpha_k = \frac{\exp{(q_{ki})}.\exp\bigg( \sum_{j \in N_i} z_{ij}. q_{kj}. \exp{-\big(||x_i-x_j||_2}\big)\bigg)}
{\sum_{k'} \exp{(q_{k'i})}.\exp\bigg(\sum_{j \in N_i} z_{ij}. q_{k'j}. \exp{-\big(||x_i-x_j||_2}\big)\bigg)} 
\label{eqn:z}
\end{equation}

When $z_{ij}=0$ (i.e., scene boundary is present within the neighborhood $[i,j]$), the update equation for $\alpha_k$ becomes similar to Equation 3 in \cite{Arandjelovic16}. When $z_{ij}=1$ (i.e., there is no scene change in the neighborhood), then it becomes similar to the temporally coherent update in Equation~\ref{eqn:assign_coh}.

The $z_{ij}$ variables can be learned during model training. We can also consider that the $z_{ij}$ values are the output of a separate pre-trained scene boundary detector.

\subsection{RNN with Temporal Coherence}
\label{sec:rnn-temporal}

\begin{figure}[hbtp]
\centering
    \includegraphics[width=0.8\columnwidth]{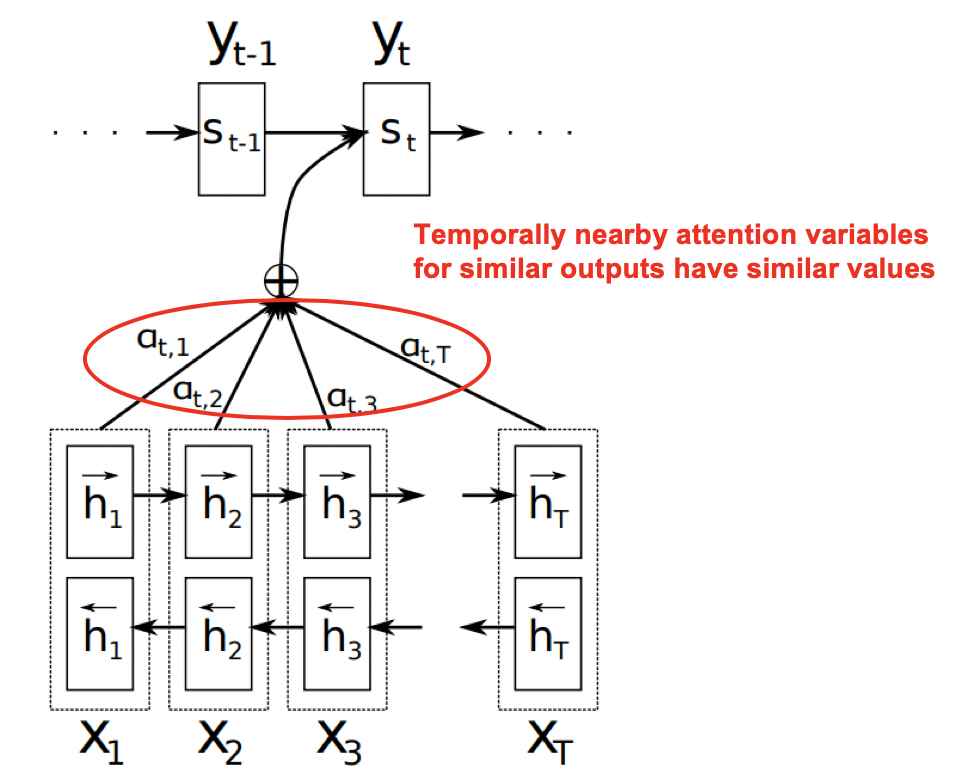}
    \caption{Temporal coherence principle applied to attention RNN model~\cite{xu2015show}.}
    \label{fig:rnn-temporal}
\end{figure}

Figure~\ref{fig:rnn-temporal} shows how the temporal coherence can be applied to the attention-based RNN model.
The principle of temporal coherence can be applied to an RNN model too, this time on the attention variables. The key idea is that if the model processing a video segment attends to a part of a sequence (e.g., a particular frame), then there is a high chance that it attends to a neighboring part of the sequence (e.g., the previous frame), where neighborhood can either is defined by temporal proximity in the video stream.

Let us consider the formulation in the original attention modeling paper~\cite{xu2015show}. In our model, the different attention values correspond to the predictions for every frame in a video segment. We consider the following form of the attention variable $\alpha_i$:
\[
\alpha_i = \frac{\exp{(e_i)}}{\sum_j \exp{(e_j)}}.
\]
Note that $e_i$ is defined as $f_{att}(a_i, h)$, where $f_{att}$ is an attention model, $h$ is the hidden state of the model. 


We propose a new formulation:
\begin{equation}
\small
\alpha_i = \frac{\exp{(e_i)}.\exp\bigg(\sum_{j \in N_i} e_j. \exp{-\big(||h_i-h_j||_2}\big)\bigg)}
{\sum_i \exp{(e_i)}.\exp\bigg(\sum_{j \in N_i} e_j. \exp{-\big(||h_i-h_j||_2}\big)\bigg)} 
\end{equation}

where $N_i$ is the neighborhood of node $i$ (say the nodes in the neighborhood $[i-L, i+L]$), while $h_i$ is the hidden layer corresponding to $e_i$.  \\

Note that using this formulation $\alpha_j$ is defined as follows:
\begin{equation}
\small
\alpha_j = \frac{\exp{(e_j)}.\exp\bigg(\sum_{i \in N_j} e_i. \exp{-\big(||h_i-h_j||_2}\big)\bigg)}
{\sum_j \exp{(e_j)}.\exp\bigg(\sum_{i \in N_j} e_i. \exp{-\big(||h_i-h_j||_2}\big)\bigg)}
\label{eq:rnn-attention}
\end{equation}

Now let us consider that $i$ and $j$ are adjacent and the only neighbors of each
other, i.e., $N_i = \{j\}$ and $N_j = \{i\}$. In this formulation, we get $\alpha_i = \alpha_j$ (as expected).

Note that if either of the conditions $N_i = \{j\}$ and $N_j = \{i\}$ are not both satisfied, i.e., either $i$ or $j$ have different neighbors, then $\alpha_i$ is not necessarily equal to $\alpha_j$ (also as expected).

\subsection{Transformer model with Temporal Coherence}
\label{sec:transformer-temporal}

Temporal coherence can be applied to Transformer models as well.

\begin{figure}[hbtp]
\centering
    \includegraphics[width=0.5\columnwidth]{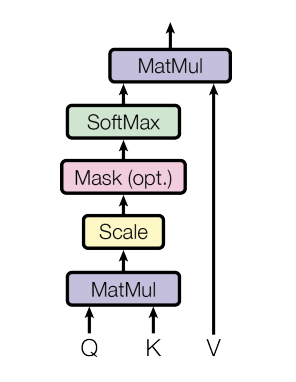}
    \caption{Scaled dot-product attention in Transformer model~\cite{Vaswani17}.}
    \label{fig:transformer}
\end{figure}

The attention formulation used for RNNs was the additive form~\cite{Bahdanau14}. In the Transformer~\cite{Vaswani17} architecture, which has comparable performance to RNNs but is more space-efficient and has faster convergence property, the multiplicative form of scaled dot-product attention is used:
\[
Attention(Q,K,V) = softmax \big(\frac{QK^T}{\sqrt{d_k}}\big).V,
\]
where Q, K and V are shown in Figure~\ref{fig:transformer}.
Note the similarity of this form to the compact additive form in Section~\ref{sec:rnn-temporal}:
\[
Attention(E,H) = softmax(E).H,
\]
where $E$ is the vector form of $e_i$ and $H$ is of $h$.
\section{Implementing Temporal Coherence using Convolutional and Pooling Layer}
\label{sec:convolution}

The temporal coherence for both the attention variables in the RNN (and Transformer) and the VLAD vectors in the NetVLAD model can be implemented efficiently using a convolutional layer --- this section outlines that connection, and shows how the temporally coherent variants of these 3 models (TC-RNN, TC-TM and TC-NetV) can be designed.

\subsection{TC-NetV: Temporally coherent NetVLAD}
\label{sec:netvlad-conv}

Let us consider the equation for NetVLAD with temporal coherence. Considering $d_{ij} = exp(-(||x_i - x_j||_2))$, we get the following form of Equation~\ref{eqn:assign_coh}:
\begin{equation}
\small
\alpha_k = \frac{\exp{(q_{ki})}.\exp\bigg(\sum_{j \in N_i} q_{kj}.d_{ij}\bigg)}
{\sum_i \exp{(q_{ki})}.\exp\bigg(\sum_{j \in N_i} q_{kj}.d_{ij}\bigg)} 
\label{eqn:d}
\end{equation}

Plugging in the value of $q_{ki} = w_k^T x_i + b_k$ and $N_i = [i-L, i+L]$, we get the equivalent form of Equation~\ref{eqn:d}:
\begin{equation}
\small
\alpha_k = \frac{\exp\bigg(w_k^T \big(\sum_{j=i-L}^{i+L} x_{j}.d_{ij} \big) + b_k\bigg)}
{\sum_i \exp\bigg(w_k^T \big(\sum_{j=i-L}^{i+L} x_{j}.d_{ij} \big) + b_k\bigg)}
\label{eqn:e}
\end{equation}

Note that the form of Equation~\ref{eqn:e} is that of a 1-dimensional convolutional layer, where the convolution is taken over a neighborhood $N_i$ of size 2L. Note that we can change that size to generate feature maps of different sizes. The final $\alpha_k$ can be an average over all these different neighborhood sizes, which can be accomplished by using a convolutional layer with average pooling.

\begin{figure}[hbtp]
\centering
    \includegraphics[width=0.8\columnwidth]{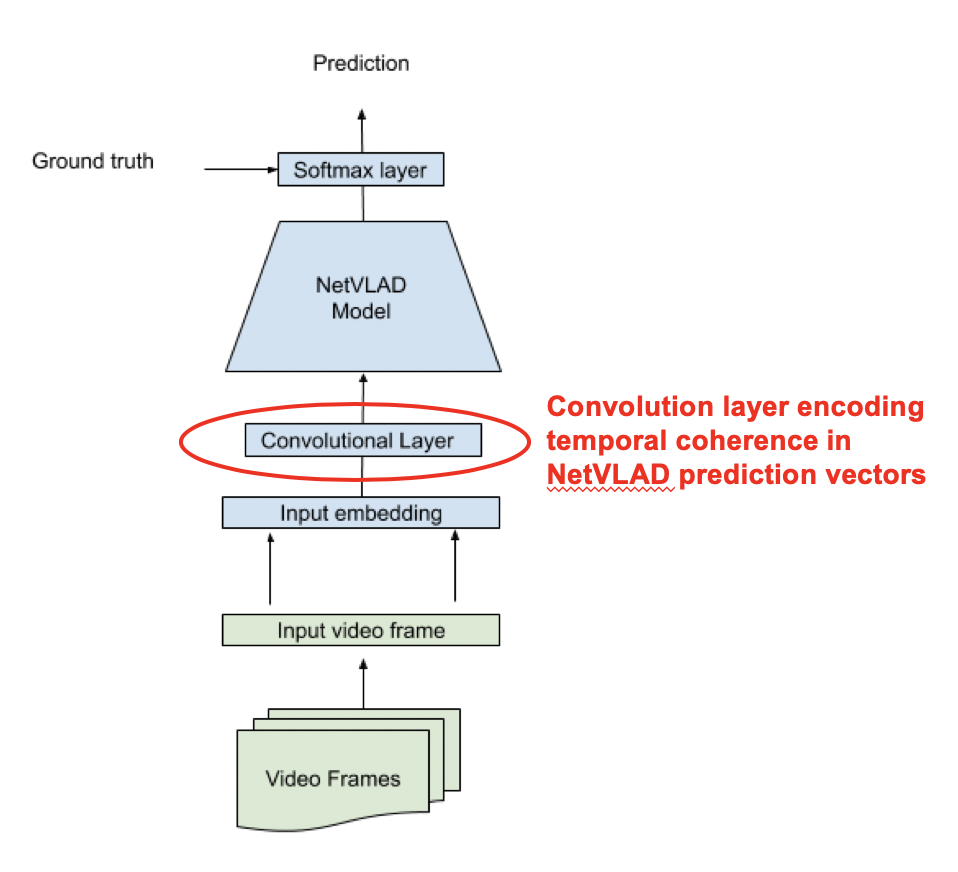}
    \caption{TC-NetV: Temporal coherent NetVLAD model.}
    \label{fig:netvlad-convolution}
\end{figure}

Thus, we have been able to show that we can efficiently encode the temporal coherence property in the NetVLAD model using a single carefully-constructed convolutional layer. Figure~\ref{fig:netvlad-convolution} shows TC-NetV, the NetVLAD model with the convolution layer to ensure temporal coherence.

\subsection{TC-RNN and TC-TM: RNN and Transformer models with Temporal Coherence Convolution}
\label{sec:rnn-conv}

We can show a similar convolution layer formulation of temporal coherence for the attention variables in the RNN model.

\subsubsection{TC-RNN: Temporally coherent RNN}

Let us consider $d'_{ij} = exp(-(|h_i - h_j||_2))$ and neighborhood to be $N_j = [j-L, j+L]$. Plugging those in, 
we get the following form of Equation~\ref{eq:rnn-attention}:
\begin{equation}
\small
\alpha_j = \frac{\exp\bigg(\sum_{i = j-L}^{j+L} e_i. d'_{ij} + e_j\bigg)}
{\sum_j \exp\bigg(\sum_{i = j-L}^{j+L} e_i. d'_{ij} + e_j\bigg)}
\end{equation}

\begin{figure}[hbtp]
\centering
    \includegraphics[width=0.8\columnwidth]{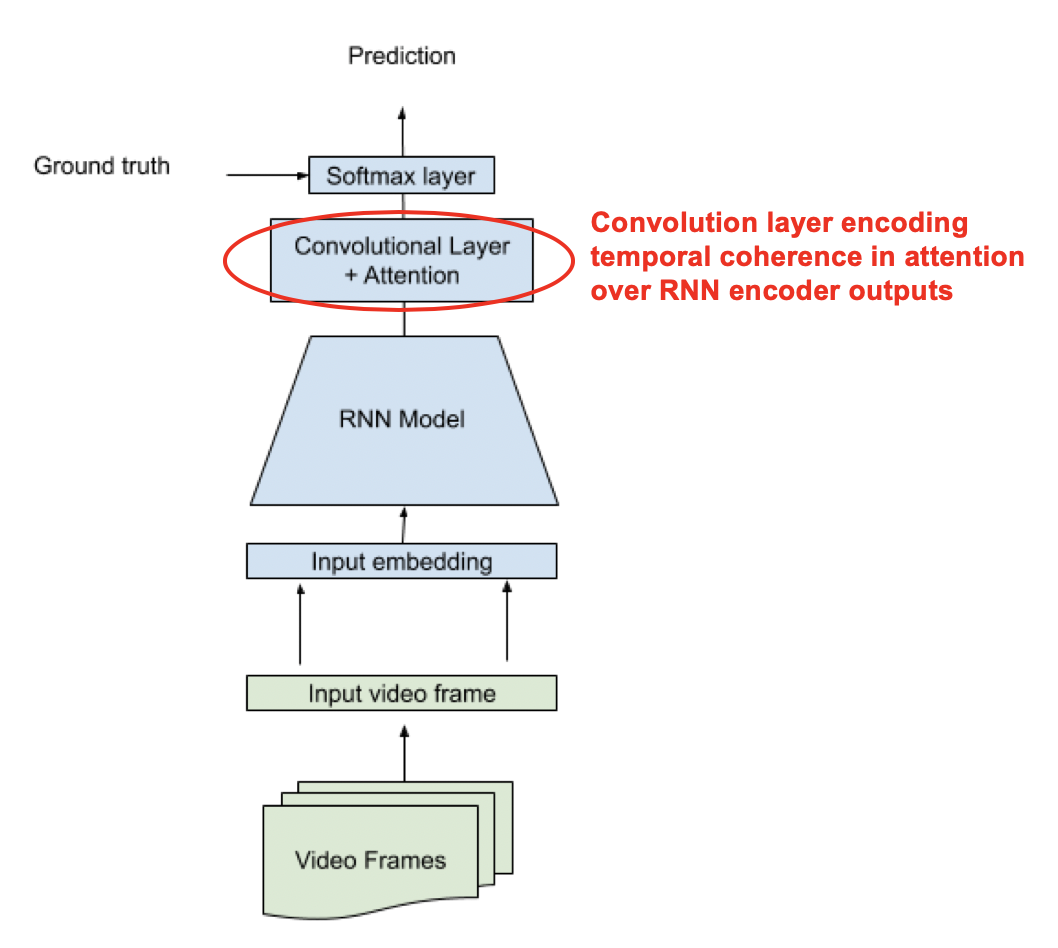}
    \caption{TC-RNN: Temporally coherent RNN model.}
    \label{fig:rnn-convolution}
\end{figure}

This is once again the form of a 1-dimensional convolutional layer, taking over a neighborhood $N_j$ of size 2L, with a final average pooling. Note that in this case the convolution layer is after the output of the encoder ($h$) in the RNN, in contrast to TC-NetV where the convolution layer was after the input. Figure~\ref{fig:rnn-convolution} shows the TC-RNN model with the convolution layer to ensure temporal coherence over attention variables defined over the encoder outputs.

\subsubsection{TC-TM: Temporally coherent Transformer}

Given the equivalence between the attention formulations of the RNN and Transformer models (as shown in Section~\ref{sec:transformer-temporal}), it can be demonstrated that temporal coherence can also be instantiated in the Transformer model using a 1-dimensional convolutional layer to get a TC-TM model (details are omitted due to lack of space). In our implementation, we use a multi-headed transformer model with 12 TC-TM models concatenated.

\subsection{TC-Ens: Temporally Coherent Ensemble model}
\label{sec:ens-conv}

\begin{figure*}[hbtp]
\centering
    \includegraphics[width=0.8\textwidth]{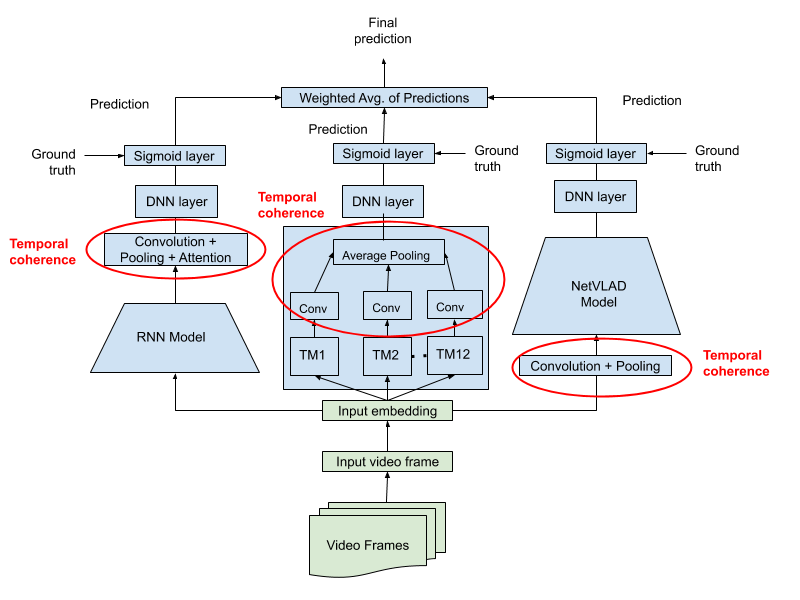}
    \caption{TC-Ens: Temporally coherent ensemble of Transformer, NetVLAD and RNN models.}
    \label{fig:ensemble-temporal}
\end{figure*}

In our experiments in the next section, we will be using an ensemble of RNN, Transformer and NetVLAD models with temporal coherence. Figure~\ref{fig:ensemble-temporal} shows the ensemble architecture TC-Ens, which will be using as one of the models in our experiments.

\section{Experimental Results}
\label{sec:experiments}
\begin{table*}[!ht]
\centering
\caption{GAP, Hit$@1$ and MAP scores (in percentages) at different hierarchical levels for different models for video categorization task on Youtube-8m. TC, NetV, TM and Base-Ens respectively denote temporal coherence, NetVLAD, transformer model and baseline ensemble. For each `TC' column the increment is shown with respect to its preceding column.}
\label{tab:youtube_model_perf}
\begin{tabular}{l||l|l|l|l|l|l|l|l|l|l}
  GAP       &    DNN     &   RNN     &   RNN-Attn   &   TC-RNN         &   NetV      &    TC-NetV        &   TM     &   TC-TM        &   Base-Ens   &   TC-Ens\\ \hline \hline
Overall     &    89.3    &   88.3   &   89.9        &   \textbf{90.7 (\textcolor{red}{+0.8})}    &   89.0    &      \textbf{90.4 (\textcolor{red}{+1.4})}    &   90.9   &   \textbf{91.7 (\textcolor{red}{+0.8})}  &   91.2  &   \textbf{\textcolor{blue}{92.1} (\textcolor{red}{+0.9})}    \\ \hline
Level 0     &    93.8    &   93.7   &   94.5        &   \textbf{94.9 (\textcolor{red}{+0.4})}    &   93.6    &      \textbf{94.4 (\textcolor{red}{+0.8})}    &   95.0   &   \textbf{95.5 (\textcolor{red}{+0.5})}  &   95.2  &   \textbf{95.7 (\textcolor{red}{+0.5})}      \\ \hline
Level 1     &    92.1    &   91.5   &   92.7        &   \textbf{93.2 (\textcolor{red}{+0.5})}    &   91.9    &      \textbf{92.9 (\textcolor{red}{+1.0})}    &   93.2   &   \textbf{93.9 (\textcolor{red}{+0.7})}  &   93.5  &   \textbf{94.2 (\textcolor{red}{+0.6})}       \\ \hline
Level 2     &    89.5    &   88.5   &   90.2        &   \textbf{91.0 (\textcolor{red}{+0.8})}    &   89.4    &      \textbf{90.8 (\textcolor{red}{+1.4})}    &   90.9   &   \textbf{91.9 (\textcolor{red}{+1.0})}  &   91.5  &   \textbf{92.3 (\textcolor{red}{+0.8})}       \\ \hline
Level 3     &    85.6    &   83.2   &   86.0        &   \textbf{87.2 (\textcolor{red}{+1.2})}    &   85.0    &      \textbf{87.3 (\textcolor{red}{+2.3})}    &   87.7   &   \textbf{88.9 (\textcolor{red}{+1.2})}  &   87.9  &   \textbf{89.3 (\textcolor{red}{+1.4})}     \\ \hline \hline
\end{tabular}
\begin{tabular}{l||l|l|l|l|l|l|l|l|l|l}
   MAP     & 	DNN      &   RNN     &   RNN-Attn 	&     TC-RNN         &   NetV   &    TC-NetV        &   TM      &   TC-TM         &   Base-Ens    &   TC-Ens\\ \hline \hline
Overall      &  86.3     &   84.2    &   86.8       &     \textbf{87.7 (\textcolor{red}{+0.9})}    &	 86.0   &     \textbf{87.9 (\textcolor{red}{+1.9})}    &   88.0    &   \textbf{89.2 (\textcolor{red}{+1.2})}  &   88.4    &   \textbf{\textcolor{blue}{89.6} (\textcolor{red}{+1.2})}          \\ \hline
Level 0      &  86.8     &   85.8    &   87.8       &     \textbf{88.3 (\textcolor{red}{+0.5})}    &   86.3   &    \textbf{87.6 (\textcolor{red}{+1.3})}    &   89.0    &   \textbf{89.9 (\textcolor{red}{+0.9})}  &   89.1    &   \textbf{90.1 (\textcolor{red}{+1.0})}        \\ \hline
Level 1 	 &  87.9     &   86.0    &   88.3       &     \textbf{89.0 (\textcolor{red}{+0.7})}   &    87.2   &    \textbf{88.9 (\textcolor{red}{+1.7})}    &   89.3    &   \textbf{90.4 (\textcolor{red}{+1.1})}  &   89.6    &   \textbf{90.8 (\textcolor{red}{+1.2})}         \\ \hline
Level 2 	 &  88.5     &   86.7    &   89.2       &     \textbf{89.9 (\textcolor{red}{+0.7})}   &    88.3   &    \textbf{90.0 (\textcolor{red}{+1.7})    } &   89.8    &  \textbf{91.0 (\textcolor{red}{+1.2})}  &   90.4    &   \textbf{91.4 (\textcolor{red}{+1.0})}         \\ \hline
Level 3 	 &  85.8     &   83.5    &   86.3       &     \textbf{87.5 (\textcolor{red}{+1.2})}   &    85.4   &    \textbf{87.8 (\textcolor{red}{+2.4})}    &   88.4    &   \textbf{89.2 (\textcolor{red}{+0.8})}  &   88.0    &   \textbf{89.3 (\textcolor{red}{+1.3})}        \\ \hline \hline
\end{tabular}

\begin{tabular}{l||l|l|l|l|l|l|l|l|l|l}
   PERR     & 	DNN      &   RNN     &   RNN-Attn 	&     TC-RNN         &   NetV   &    TC-NetV        &   TM    &   TC-TM        &   Base-Ens  &   TC-Ens\\ \hline \hline
Overall      &  86.2     &   85.3    &   86.9       &     \textbf{87.6 (\textcolor{red}{+0.7})}    &   85.8  &      \textbf{87.1 (\textcolor{red}{+1.3})}    &   88.1  &   \textbf{88.8 (\textcolor{red}{+0.7})}  &   88.4 &   \textbf{\textcolor{blue}{89.1} (\textcolor{red}{+0.7})}        \\ \hline
Level 0      &  91.4     &   91.1    &   92.1       &     \textbf{92.5 (\textcolor{red}{+0.4})}    &   91.1   &    \textbf{92.0 (\textcolor{red}{+0.9})}    &   92.8  &   \textbf{93.3 (\textcolor{red}{+0.5})}  &   93.1 &   \textbf{93.5 (\textcolor{red}{+0.4})}       \\ \hline
Level 1 	 &  89.9     &   89.3    &   90.5       &     \textbf{91.0 (\textcolor{red}{+0.5})}    &   89.6   &    \textbf{90.7 (\textcolor{red}{+1.1})}    &   91.3  &   \textbf{91.9 (\textcolor{red}{+0.6})}  &   91.6 &   \textbf{92.2 (\textcolor{red}{+0.6})}        \\ \hline
Level 2 	 &  85.6     &   84.1    &   86.0       &     \textbf{86.7 (\textcolor{red}{+0.7})}    &   85.3   &    \textbf{86.8 (\textcolor{red}{+1.5})}    &   87.1  &   \textbf{87.9 (\textcolor{red}{+0.8})}  &   87.6 &   \textbf{88.4 (\textcolor{red}{+0.8})}          \\ \hline
Level 3 	 &  81.9     &   79.6    &   81.8       &     \textbf{83.2 (\textcolor{red}{+1.4})}    &   81.6   &    \textbf{83.7 (\textcolor{red}{+2.1})}    &   83.6  &   \textbf{84.9 (\textcolor{red}{+1.3})}  &   84.4 &   \textbf{85.6 (\textcolor{red}{+1.2})}         \\ \hline \hline
\end{tabular}

\begin{tabular}{l||l|l|l|l|l|l|l|l|l|l}
   Hit@1     & 	DNN      &   RNN     &   RNN-Attn 	&     TC-RNN         &   NetV   &    TC-NetV        &   TM    &   TC-TM        &   Base-Ens  &   TC-Ens\\ \hline \hline
Overall      &  93.1     &   92.7    &   93.5       &     \textbf{93.8 (\textcolor{red}{+0.3})}    &	 92.9  &      \textbf{93.6 (\textcolor{red}{+0.7})}    &   94.2  &   \textbf{94.5 (\textcolor{red}{+0.3})}  &   94.4 &   \textbf{\textcolor{blue}{94.8} (\textcolor{red}{+0.4})}        \\ \hline
Level 0      &  93.2     &   92.9    &   93.7       &     \textbf{94.0 (\textcolor{red}{+0.3})}    &   93.0   &    \textbf{93.7 (\textcolor{red}{+0.7})}    &   94.3  &   \textbf{94.7 (\textcolor{red}{+0.4})}  &   94.6 &   \textbf{94.9 (\textcolor{red}{+0.3})}       \\ \hline
Level 1 	 &  92.4     &   91.8    &   92.8       &     \textbf{93.2 (\textcolor{red}{+0.4})}    &   92.2   &    \textbf{93.1 (\textcolor{red}{+0.9})}    &   93.5  &   \textbf{93.9 (\textcolor{red}{+0.4})}  &   93.7 &   \textbf{94.2 (\textcolor{red}{+0.7})}        \\ \hline
Level 2 	 &  89.4     &   88.0    &   89.9       &     \textbf{90.4 (\textcolor{red}{+0.5})}    &   89.3   &    \textbf{90.4 (\textcolor{red}{+1.1})}    &   90.5  &   \textbf{91.2 (\textcolor{red}{+0.7})}  &   90.9 &   \textbf{91.5 (\textcolor{red}{+0.6})}          \\ \hline
Level 3 	 &  86.3     &   84.3    &   86.5       &     \textbf{87.2 (\textcolor{red}{+0.7})}    &   86.0   &    \textbf{87.9 (\textcolor{red}{+1.9})}    &   87.5  &   \textbf{88.2 (\textcolor{red}{+0.7})}  &   87.9 &   \textbf{89.0 (\textcolor{red}{+1.1})}         \\ \hline \hline
\end{tabular}
\end{table*}
In this section, we showcase the improvement in performance of video categorization by using temporal coherence. We first present details of Youtube-8M data set used for our experiments and the hierarchical taxonomy we created for enhancing the data set. We then explain the experimental setup, followed by the results and its analysis. 

\subsection{Dataset and Hierarchical Taxonomy Creation} \label{sec:data}
For our experiments we use Youtube-8M dataset consisting of 5 million Youtube videos, with lengths clipped to 5 minutes. Each video is tagged using Youtube annotation system that uses video title, description and tags. For efficient storage, the dataset is pre-processed using ImageNet and PCA and each frame in the video is represented by a 1024 dimensional video descriptor and 128 dimensional audio descriptor. Further, both video and audio frames are sampled at 1 frame per second.

The tags generated by the annotation system span a vocabulary of ~3800 labels. Many of these labels are quite similar. For example, there are 7 tags corresponding to the game ``The Sims'' including ``The Sims'', ``The Sims (video game)'', ``The Sims Freeplay'', ``The Sims 2'', ``The Sims 3'', ``The Sims 3 (Pets)'', ``The Sims 4''. Such flat label space poses two problems. Firstly, for a video recognition tool, identifying these 7 variants is less critical than identifying the overall game. Secondly, if the model identifies the game ``The Sims'' as ``The Sims (video game)'', it will be penalized the same as if the prediction was ``Food'' which is incorrect as the former label is much closer to the ground truth. 

\begin{figure}[hbtp]
\centering
    \includegraphics[width=\columnwidth]{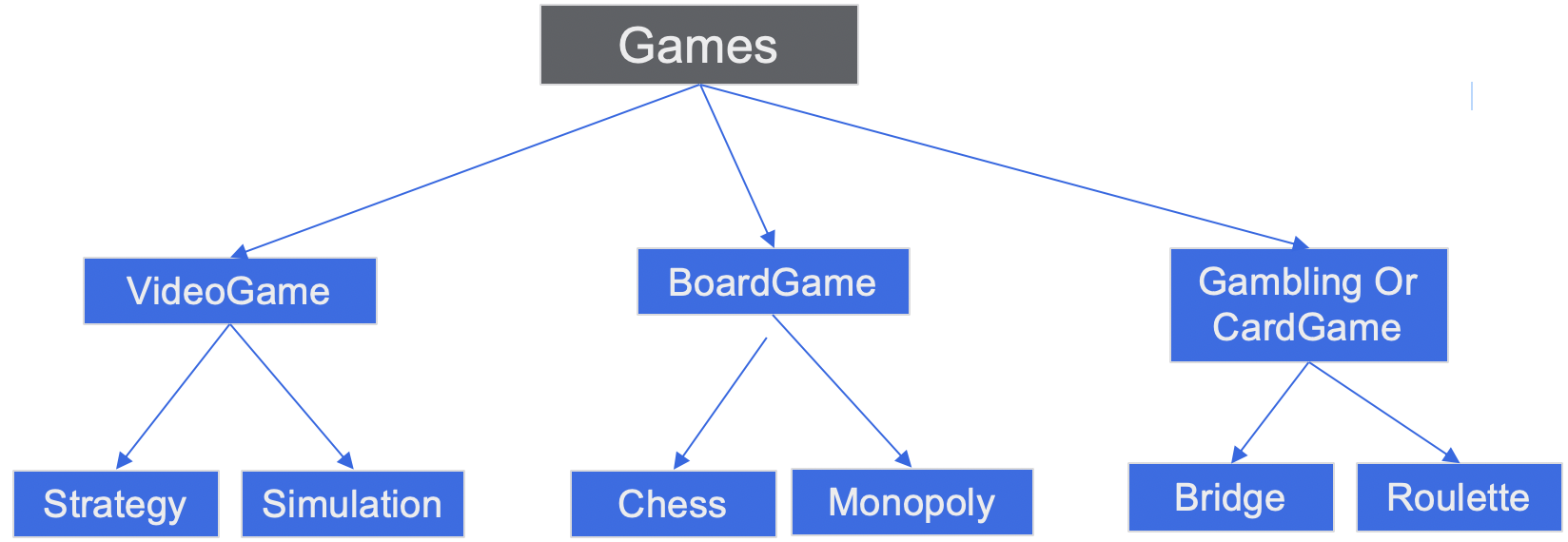}
    \caption{Hierarchical taxonomy for the top label ``Game''.}
    \label{fig:hierarchy}
\vspace{-10pt}
\end{figure}


To overcome this, we created a hierarchical taxonomy of labels. As an example, in Figure~\ref{fig:hierarchy}, we show the top 3 levels of the taxonomy created for ``Games''. The taxonomy was created and verified by 3 engineers. Branches of a label were created to ensure comprehensiveness and mutual exclusion. The final hierarchical taxonomy contains 18 top-level labels including ``MovieOrTV'', ``Science'', ``Home\&Garden'', ``Music'', ``Art'', ``Game'', ``Fitness'', ``Beauty'', ``Toy'', ``Fashion'', ``Books'', ``Transport'', ``PublicServices'', ``Travel'', ``Electronics'', ``Animals'', ``Sports'' and ``Food''. The taxonomy has a total of 640 nodes and a maximum of 4 levels.

We created an annotation task to map the 3800 labels to the 640 nodes in our taxonomy. We asked 5-7 annotators to look at the videos for each Youtube-8M label and assign it to a node in our taxonomy. After the completion of the task, three engineers manually went through each of the assignments and removed inconsistencies. 
\subsection{Experimental set-up and Hyperparameters} \label{sec:architecture}
We performed all our experiments on a 64 bit Ubuntu 18.0 LTS system with Intel (R) Core (TM) i9-9820X CPU with 10 CPU cores, 3.30 GHz CPU clock frequency, 64 GB RAM, and two Nvidia GeForce RTX 2080 Ti, each with 12 GB memory.

For the task of multi-label video categorization, we evaluate our model using four metrics: (i) Global Average Precision, which computes the area under curve of the overall precision-recall curve, (ii) Mean Average Precision, which computes a per class area under curve of precision recall curve and averages them, (iii) Precision at Equal Recall Rate (PERR), which computes the mean precision up to the number of ground truth labels in each class, and (iii) Hit@1, which is the fraction of the test samples that contain a ground truth label in the topmost prediction.

For all the experiments, we train our model on 4 million videos, use 64k videos to tune model hyperparameters and early stopping, and perform 5 test runs with 128k videos each, randomly sampled from the remain 934k videos. We report the mean of the 5 test runs. Standard deviation in each case was less than 0.03.

For comparison, we run a vanilla deep neural network (referred as DNN hereafter) on the aggregated video level features for which the aggregation was done by applying LSTM on the frame level features. We ran a grid search on the DNN architecture to identify the best performing architecture and used it on the features generated by RNN, NetVLAD and Transformer models. For NetVLAD and RNN, we used the best architectures identified by Next top GB model \cite{skalic2018building}, which won the Youtube-8M challenge. For NetVLAD, we use 192 clusters. For RNN, we used 512 hidden GRU units. We used the same architecture for TC-NetV and TC-RNN as well for fair comparison. Our model has two new hyperparameters, kernel size and number of feature maps. We do a grid search on them in the set  \{5, 9, 13, 17, 21\} and \{1, 2, 4, 8, 16\} respectively. We got best results for kernel size 5 and feature map 4. We used Adam optimizer for training with initial learning rate of 0.0002 and batch size of 128. We padded 0 frames at the end of each video to get 300 frames per video. Each model was trained for a maximum of 10 epochs.
\subsection{Results}
\label{sec:results}

Mean values of 5 test runs of the evaluation metrics are presented in Table~\ref{tab:youtube_model_perf}. We show results on temporal coherence for each individual baseline model. Further, we report results of the ensemble of all baseline models and all temporal coherent model. For each metric, we have (a) an overall result that considers the taxonomy as a flat hierarchy, and (b) level-based results which only considers predictions for a given level in the label hierarchy.

\begin{table*}[hbtp]
\centering
\caption{Training time (for 1 epoch) and model size of the models.}
\label{tab:youtube_model_time}
\begin{tabular}{l|l|l|l|l|l|l|l|l}
                    &    DNN      &  RNN      &   RNN-Attn     &   TC-RNN   &   NetV        &   TC-NetV  &   TM      & TC-TM\\ \hline \hline
Training Time       &    40 min   &  55 min   &   65 min       &   67.5 min &   130 min     &   133 min  &   102 min & 110 min           \\ \hline
Model Size          &    4.7 MB   &  14.2 MB  &   14.2 MB      &   14.2 MB  &   595 MB      &   595 MB   &   52 MB   & 52 MB   \\ \hline \hline
\end{tabular}
\end{table*}



First, let us observe the overall scores. We see that all the models show performance boost with the temporal coherence, with TC-Ens having the best performance overall. The gain in performance for RNN and Transformers is around 0.9 for GAP and MAP, which is significant. NetVLAD models gain even more with a gain of 1.4 and 1.7 for GAP and MAP respectively. This is intuitive as RNN and Transformers capture temporal patterns but NetVLAD purely clusters based on distance in the feature space, regardless of time. Thus, adding temporal coherence adds more to the modeling capability of NetVLAD. Further, we observe that the gain for MAP is consistently more compared to gain in GAP. As effectively MAP is weighting underrepresented classes more than GAP, this suggests having temporal coherence is even more beneficial for classes with lesser examples. This is validated by the lower increase in Hit@1 values. As the top prediction can be a generic label, learning patterns of classes with ample examples is sufficient to get a high Hit@1 value.

We next dive into level-based metric values. Overall, we see that for all models, performance decreases as we go deeper in the taxonomy as illustrated by GAP and Hit@ 1. There are two reasons for this. Firstly, the difference between videos in a lower level is more fine-grained and harder to capture. Secondly, the number of training examples are lesser for deeper levels, since a shallower level is a super set of all levels under it.  Interestingly, we see that with temporal coherence the performance decrease in accuracy as we consider increasing levels is lesser than models without coherence, strengthening our hypothesis that temporal coherence enables the model to handle more difficult classification problems (due to effective regularization) and makes the model more robust to smaller sample size. We thus observe a greater performance improvement over the  baseline models for lower levels.

We present the amount of training time per epoch and the model size for each model in Table~\ref{tab:youtube_model_time}. We see that adding temporal coherence to a model adds 2-8 min in training time per epoch, which is insignificant. Further, as the coherence only adds a few parameters, the model size increases with new hyper-parameters in KBs and thus remains practically same.

\subsection{Analysis}
We will now show what is being learned by the temporal coherence layer and why it is improving over the baseline. We begin with qualitative analysis comparing examples of videos and predictions made by our model. We then visualize the kernels learned by temporal coherence layers in TC-RNN, TC-NetV and TC-TM. We then visualize the attention weights for an example video before and after temporal coherence, showcasing the utility of the layer. 

\subsubsection{Qualitative Analysis}\label{sec:qual}
\begin{figure*}[!hbtp]
\centering
    \includegraphics[width=0.95\textwidth]{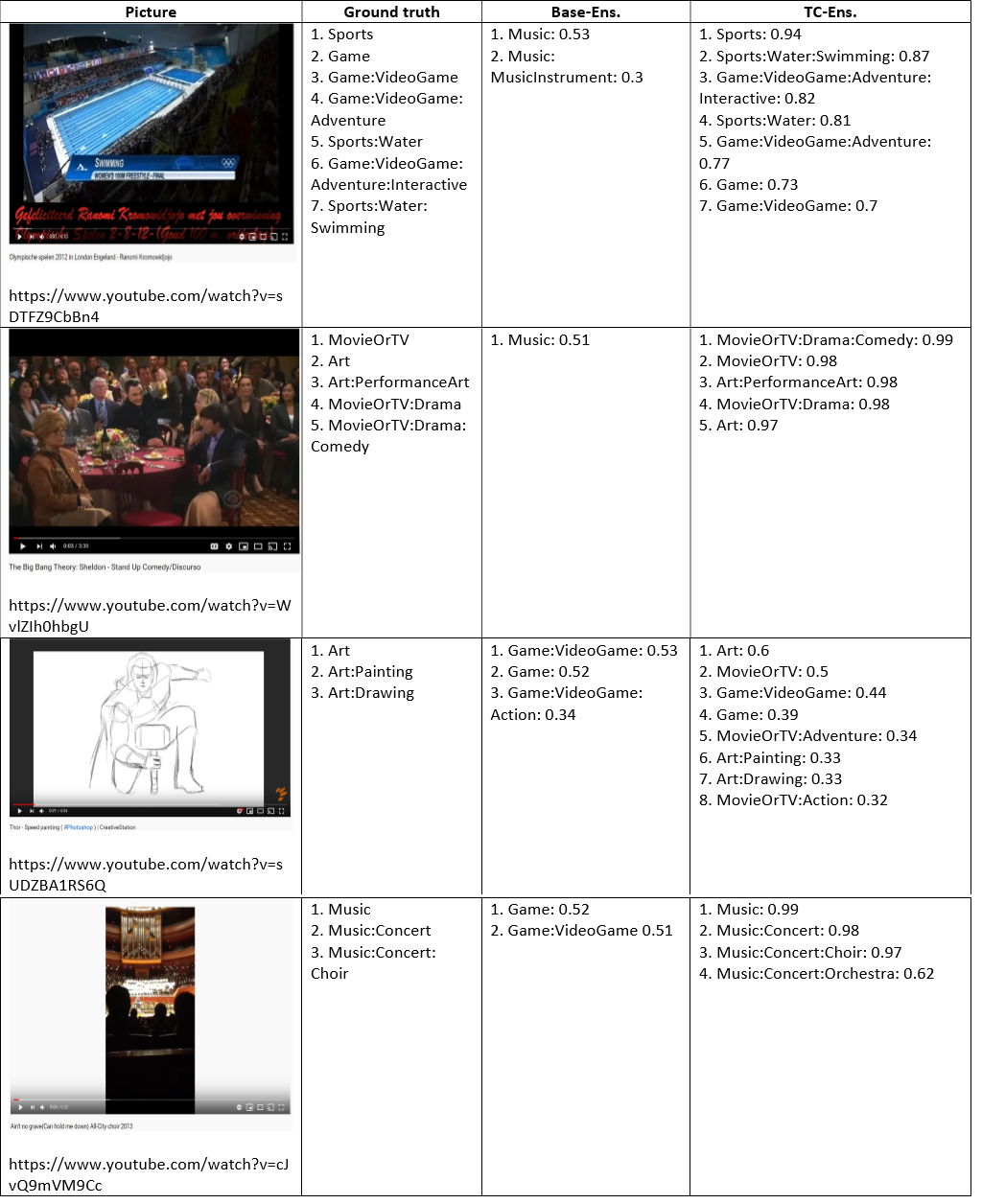}
     \caption{Qualitative analysis for some of the videos in the YouTube 8M dataset}
     \label{fig:qualitative}
\end{figure*}

In Figure~\ref{fig:qualitative} we compare the performances of the baseline ensemble model with TC-Ensemble model on a few samples from Youtube-8M dataset. For each video its ground truth labels consist of labels from all the hierarchies. For example, the first entry in the table is a video of a swimming competition. In our hierarchical taxonomy the label 'swimming' occurs at third level and is denoted as 'Sports:Water:Swimming'. This implies that the video also gets assigned 'Sports' and 'Sports:Water' as ground truth labels along with 'Sports:Water:Swimming'. For the first video, we see that baseline ensemble model fails to identify the correct ground truth labels and classifies it as 'Music' with a probability of 0.53. With temporal coherence however, we correctly classify it as a swimming video with a high probability. Similarly for the videos in rows 2 and 4 while the baseline ensemble model fails to identify their classes correctly, TC-Ensemble model classifies them correctly with high probabilities. The video in third row is an interesting case where TC-Ensemble assigns some probability of it being a video-game while actually its an 'Art' video where a person is drawing the fictional character 'Thor' in photo-shop. Nonetheless, it still correctly identifies it as an 'Art' video with a probability of 0.6 while baseline ensemble model misclassifies it completely. Also the probability with which TC-Ensemble classifies it as a video game is $< 0.5$ showing it's more confident that it's an 'Art' video than it's 'video-game'. We assume the label video-game appeared with TC-ensemble (and baseline ensemble) model because 'Thor' being a popular character in video games confuses it into believing its a video game. Also, upon watching the video we observe that the graphics from photo-shop make it seem like its a frame from a video-game. The consistent superior performance of the TC-Ensemble model over baseline model shows that incorporating temporal coherence in the models aids in video classification by taking context into account. 

\subsubsection{Analyzing kernels}\label{sec:kernel}
\begin{figure}[!hbtp]
\centering
    \includegraphics[width=0.95\columnwidth]{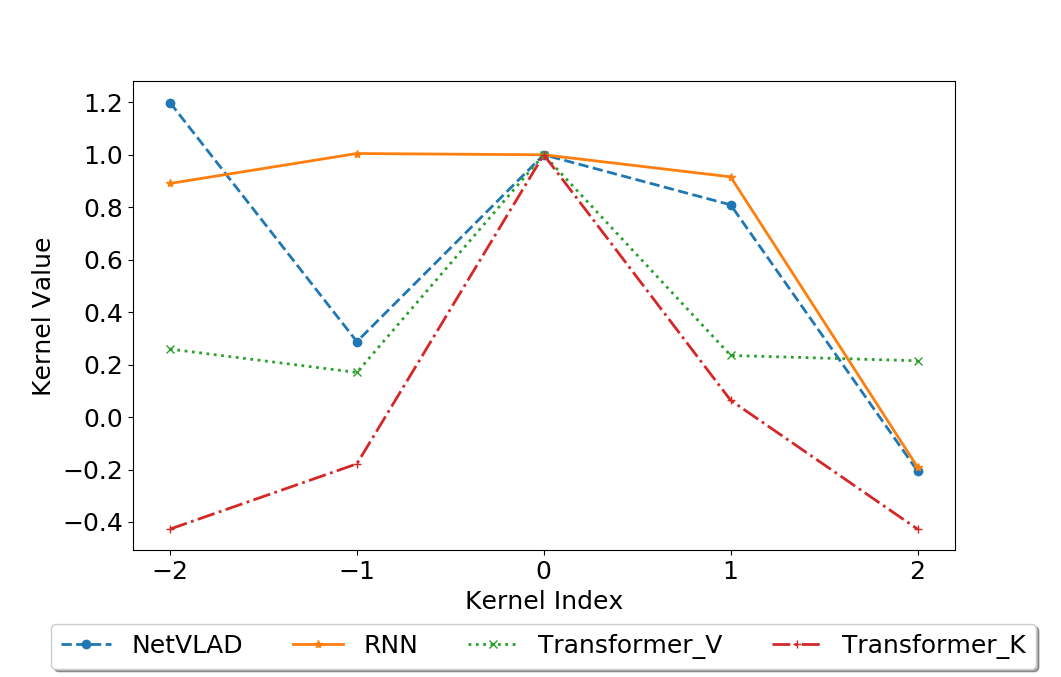}
    \caption{Averaged kernel over four feature maps of our temporal convolution layer for all the models. Kernel size is 5. Transformer\_V refers to the kernel for TC layer on top of the parameter V of transformer and Transformer\_K is for the parameter K of transformer. Note that the kernel values have been normalized so that it can be plotted together.}
    \label{fig:kernel_rnn}
\end{figure}

The temporal convolution layer used to implement temporal coherence consists of temporal kernels. Kernel weights represent the amount of importance given to the corresponding time step. Figure~\ref{fig:kernel_rnn} illustrates kernel weights for TC-RNN, TC-NetV and TC-TM for kernels of size 5. We see that highest weight is given to the current time step and neighboring time steps are given lower weight as the distance increases. In general, we observe a bell curve with peak at 0. 

\subsubsection{Visualizing attention with and without Temporal convolution layer}\label{sec:vis}
\begin{figure}[!hbtp]
\centering
\begin{tabular}
{@{\hspace{-0.2cm}}c@{\hspace{0.0cm}}}
{\includegraphics[height=4.5cm, width=4.5cm]{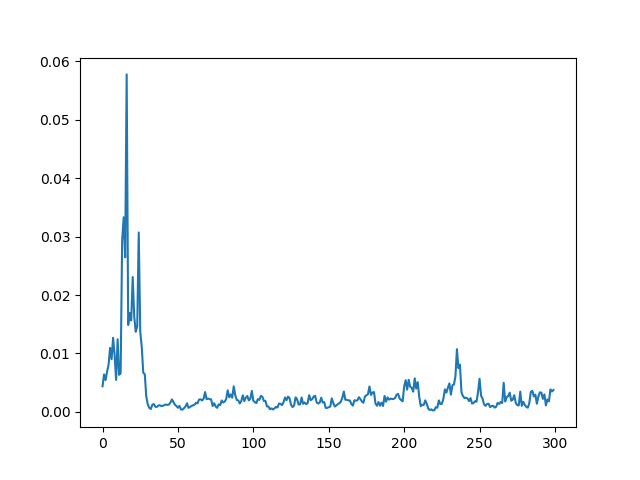}} 
{\includegraphics[height=4.5cm, width=4.5cm]{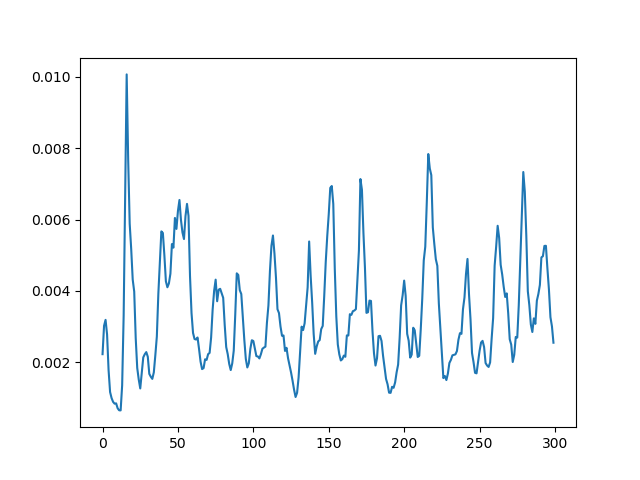}}
\end{tabular}
\vspace{-4mm}
\caption{Attention weights without (left) and with (right) our temporal coherence. Note how without temporal coherence attention is concentrated on the initial few frames while TC-RNN attends to frames throughout the time.}
\label{fig:hyper}
\vspace{-2mm}
\end{figure}

We now look at the attention weights with and without the temporal coherence layer in RNN-Attn model. Figure~\ref{fig:hyper} shows attention weights for the time steps for a sample video. The video~\footnote{\url{https://www.youtube.com/watch?v=a3RexCQS33Q}} begins with a dark stage at around 15 seconds some singers appear. The baseline RNN model gives a high attention for that time step. With temporal coherence, the model gives a much more uniform attention due to which it is able to capture the activity of singing making correct prediction. The RNN model predicts ``Game'' for this video whereas TC-RNN predicts correctly ``Music-Concert''.
\section{Related Work}
\label{sec:related}

Multimodal ML models have been developed for video analysis for a variety of tasks, e.g., video indexing~\cite{snoek2005multimodal}, video search~\cite{francis2019fusion}, emotion detection~\cite{ortega2019multimodal}, virtual reality scene construction~\cite{wang2019virtual}, collaborative learning~\cite{chua2019edubrowser}, 
learning multi-modal representations of contact-rich data types~\cite{lee2019making}, task-agnostic video-linguistic representations~\cite{lu2019vilbert}, explanation~\cite{SelvarajuL2019}, incremental learning~\cite{Li2019,zhang2020rec},  visual dialog~\cite{zhang2019}, and even cyber-security applications like automated categorization of darkweb sites having multimodal (video, text, etc.) content~\cite{ghosh2017}.
In this paper, we focus on the multimodal content recommendation problem. 

In the context of multimodal modeling, different types of regularization approaches have been used, e.g., spatio-temporal regularization~\cite{Kundu2016CVPR}, self-regularization~\cite{lei2019fully}, etc. 
Temporal information has been studied in videos by different contexts, to understand how much temporal signals matter in video analysis~\cite{Huang2018CVPR}.
Previous work also explores the use of temporal context using a fully connect neural network
algorithm that introduces extra neurons~\cite{BeckerHinton92}. Temporal consistency has been studied in the context of detecting adversarial attacks in videos~\cite{Xiao2019ICCV}, or coming up with unsupervised embeddings of video frames~\cite{Wang2015}. Temporal coherence has also been used on unlabeled data to act as a regularizer for supervised tasks~\cite{MobahiCW09} --- in that case the model considered was CNN, while we have applied the principle of temporal coherence to multiple models (attention RNN, NetVLAD, Transformer), and also evaluated the performance of the temporally coherent versions of these models on a much larger-scale dataset compared to~\cite{MobahiCW09}. We also show how convolutional layers can be effectively used to enforce temporal coherence --- large scale deep neural networks for video classification have used convolutional layers effectively~\cite{Karpathy2014CVPR}, but not specifically to utilize them for temporal coherence enforcement. Note that we observe using prior initialization improves model performance, as has been seen too in other models~\cite{ghosh2016,papai2012}. We get the best performance when using temporal prior in the convolutional layers.

We ran our experiments on a variant of the YouTube-8M dataset~\cite{abu2016youtube}, which we curated internally --- our approaches were able to beat this benchmark. The  class hierarchy has been used effectively to define a hierarchical cross-entropy loss function~\cite{bertinetto2019making}, which we will explore in future work. Note that such feature-rich hierarchies have also been developed for accurate object detection and semantic segmentation~\cite{Girshick2014CVPR}.

\section{Conclusions and Future Work}
\label{sec:conclusions}

In this paper, we introduced the novel concept of {\it temporal coherence} for multi-modal video processing for various models like Transformers, attention RNN and NetVLAD. We theoretically show how temporal coherence can be formulated as a convolution layer with pooling for these models. Our experiments demonstrate the state-of-the-art performance that can be achieved in the task of hierarchical video categorization on the Youtube-8M dataset with our proposed TC-TM, TC-NetV, TC-RNN models, as well as the TC-Ens model (which is an ensemble of the 3 individual TC models with temporal coherence). Our temporally coherent models outperformed the baselines models consistently with statistical significance, without significant increase in training time and model size. We also see that adding temporal coherence makes the model more effectively handle the difficult problem of handling fine-grained classification towards the leaf-levels of the category hierarchy. In the future, we would like to try other ways of multimodal fusion, design a hierarchical loss function and extend this concept for other tasks and datasets in video processing (e.g., video captioning, activity recognition from videos).

\bibliographystyle{icml2020}
\bibliography{main}

\end{document}